\documentclass[pdflatex,sn-basic,Numbered]{sn-jnl}%,lineno,iicol]{sn-jnl}

\usepackage{graphicx}%
\usepackage{multirow}%
\usepackage{amsmath,amssymb,amsfonts}%
\usepackage{amsthm}%
\usepackage{mathrsfs}%
\usepackage[title]{appendix}%
\usepackage{xcolor}%
\usepackage{textcomp}%
\usepackage{manyfoot}%
\usepackage{booktabs}%
\usepackage{algorithm}%
\usepackage{algorithmicx}%
\usepackage{algpseudocode}%
\usepackage{listings}%

\usepackage{natbib}%追加
\usepackage{anyfontsize}

\begin{document}

\title[Article Title]{Generating In-store Customer Journeys from Scratch with GPT Architectures}


\author*[1]{\fnm{Taizo} \sur{Horikomi}}\email{horikomi@nii.ac.jp}

\author[2,1]{\fnm{Takayuki} \sur{Mizuno}}\email{mizuno@nii.ac.jp}

\affil*[1]{\orgname{The Graduate University for Advanced Studies, SOKENDAI}, \orgaddress{\street{Shonan Village}, \city{Hayama}, \postcode{240-0193}, \state{Kanagawa}, \country{Japan}}}

\affil[2]{\orgname{National Institute of Informatics}, \orgaddress{\street{2-1-2 Hitotsubashi}, \city{Chiyoda}, \postcode{101-0003}, \state{Tokyo}, \country{Japan}}}


\abstract{We propose a method that can generate customer trajectories and purchasing behaviors in retail stores simultaneously using Transformer-based deep learning structure. Utilizing customer trajectory data, layout diagrams, and retail scanner data obtained from a retail store, we trained a GPT-2 architecture from scratch to generate indoor trajectories and purchase actions. Additionally, we explored the effectiveness of fine-tuning the pre-trained model with data from another store. Results demonstrate that our method reproduces in-store trajectories and purchase behaviors more accurately than LSTM and SVM models, with fine-tuning significantly reducing the required training data.}




\maketitle

\section{Introduction}\label{sec1}
Understanding customer behavior indoors is crucial for various industries, including retail, shopping malls, hospitals, banks, train stations, building management companies, and public facility operators. This understanding enables stakeholders to discern customer characteristics, improve shopping experiences, optimize indoor store layouts, and ultimately, enhance profitability. Furthermore, there also is a pure scientific objective to elucidate universal laws or dynamics regarding human behavior indoors. Consequently, extensive research has been conducted to understand the movement of people indoors. The approaches of previous studies can be broadly classified into three categories.

The first approach involves physical methodologies. Wang et al. used a Markov model to predict indoor trajectories \cite{bib1}. However, the aim of this study is to generate data with properties similar to real-world data, which cannot be achieved with physical models. The second approach utilizes agent-based modeling techniques. Terano et al. devised a method called ABISS, which simulates indoor customer behavior using agent-based techniques \cite{bib2}. Goto et al. replicated in-store customer journeys and purchasing behavior within stores using agent-based simulation \cite{bib3}. A weakness of agent-based models is that they rely on known customer behaviors, making it impossible to incorporate unknown customer behaviors. Therefore, it does not align with our goal of creating data that closely resembles real-world scenarios. The third approach involves the use of machine learning techniques.  Das et al. use RNNs to predict the trajectory of people inside a building \cite{bib4}. Like Wang et al., who combined LSTM and Markov Chain, there are studies that combine some of these approaches in a hybrid manner \cite{bib5}. However, because these methods have shorter-term memory compared to Transformer-based models, they cannot generate the entire in-store customer journey. Additionally, while some methods, such as LSTM, predict position information directly using $x$ and $y$ Cartesian coordinate systems \cite{bib13}, there are also studies that focus on handling only the movement patterns between predefined blocks or areas \cite{bib14}.

On the other hand, in the field of outdoor trajectory generation, methods utilizing Transformer structures have been proposed. Mizuno, Fujimoto, and Ishikawa devised a novel approach by introducing a unique hierarchical location token that represents a geographic location \cite{bib6}. They represented individual daily trajectories around Kyoto-city using these tokens and trained a model from scratch using the GPT-2 architecture to generate locations. The model has succeeded in generating trajectories that start from home, visit some destinations, and return home again, thanks to its long memory length. Based on this research, Horikomi, Fujimoto, Ishikawa, and Mizuno demonstrated the generation of not only spatial information but also spatiotemporal information, specifically, the simultaneous generation of locations and time intervals between locations \cite{bib7}. It can be said that these methods with long term memory are also considered suitable for generating the entire in-store customer journey. Furthermore, considering the potential for the simultaneous generation of locations and time intervals, it could be possible to simultaneously generate not only trajectories but also purchasing behaviors in one’s in-store customer journey.

In this study, we aim to apply the model proposed by Mizuno, Fujimoto, and Ishikawa to smaller-scale settings, specifically retail stores, and to extend it to enable the simultaneous generation of not only customer trajectories but also purchase behaviors by learning in-store customer journeys with GPT architecture from scratch. While there have been studies predicting purchase occurrence solely based on the time spent in front of shelves \cite{bib15,bib16}, there are no studies that generate purchase behavior simultaneously with customer trajectories. By employing our proposed model, it becomes feasible to generate realistic individual trajectories and purchasing behaviors for each person, while also satisfying macro-level customer statistics such as local in-store traffic flow and zone-specific purchase counts within the store. Moreover, we seek to demonstrate the feasibility of utilizing a pre-trained model, trained with data from a particular store, to extend its applicability to different stores through fine-tuning even with limited data. Acquiring in-store trajectories is costly, so being able to generate in-store customer journeys from limited data would likely lead to significant cost reductions.

Autoregressive transformers can be regarded as a highly accurate Markov chain. By applying attention mechanisms, they identify which conditional probabilities are important from the data, resulting in highly precise predictions. This allows us to build a model that predicts customer behavior with high accuracy. Using this model to generate human behaviors and inductively examine where attention is focused can reveal the dynamics of human actions. Like in other scientific realms \cite{bib18}, this shift from hypothesis-driven to data-driven machine learning for discovering important points is the key contribution of this paper.

This paper is structured as follows: In Section 2, we describe the data used in this study. In Section 3, we provide a detailed description of the methodology in which we utilize Transformer model based on natural language generation techniques . After discussing the method of textualizing indoor location information, incorporating purchasing behavior into the model, explaining the learning process of the model, and describing the generation method using the trained model, we then explain the process of fine-tuning using data from a different store. In Section 4, we present the results. We first discuss the accuracy of the pre-training model created by training the GPT-2 architecture from scratch using the data from one of the given stores. We compare the generated trajectories and purchase behaviors with those generated using LSTM and SVM models. Then, we compare the accuracy of the generated results based on the difference in the number of training data when fine-tuning the pre-trained model with data from another store. Finally, we provide the conclusion, highlighting the contributions of this study and outlining future prospects in Section 5.

\section{Data}\label{sec2}

The study utilized data provided by TSURUHA HOLDINGS, INC., a major retail chain with over 2,000 stores nationwide in Japan. The data consists of one year's worth of customer trajectory data, layout diagrams, and retail scanner data from 2021 obtained from two stores located in Tokyo metropolitan area.  Store A is located in Kanagawa Prefecture and is a suburban store with a sales floor area of approximately 900 square meters. Store B is located in Tokyo and is an urban store with a sales floor area of approximately 500 square meters. 

The customer trajectory data was obtained using the Quuppa system \cite{bib8}, which captures indoor trajectories with a precision of 50 centimeters at 5-second intervals by installing receivers on the ceiling and transmitters on shopping baskets. It's important to note that data for individuals not carrying baskets was not captured due to limitations in data acquisition methods. 

“Layout diagram” refers to the floor plan of a store, indicating which product categories are placed in which zones within the store. Store A consists of 61 zones, while Store B consists of 41 zones.

The retail scanner data contains sales data recorded at the checkout register when products are purchased, and it includes information about the items sold such as sales date and time, unique product code, product name, sales price, sales quantity, and product category. 

The scanner data and the customer trajectory data are matched using timestamps from checkout transactions at the checkout registers. By comparing the category information from the scanner data with the layout diagram, it is possible to determine where the products were placed.

\section{Methods}\label{sec3}

\subsection{Building the Pre-trained Model}\label{subsec1}

In this study, we train the GPT-2 small architecture, a natural language-based Transformer model which consists of 12 layers of Transformer encoder \cite{bib9}, from scratch by representing the trajectory, which is expressed in a temporal sequence of locations, as text-like data. Using the data from Store A, we trained the architecture and constructed a pre-trained model. The process involved three steps: textualization of indoor location information, modeling of purchasing behavior, and tokenization and training.

Models using the GPT architecture, such as GPT, GPT-2, GPT-3, and GPT NeoX, essentially share the same architecture, with the primary difference being their size. The number of parameters in a Transformer model and the number of tokens in the training data should be approximately equal for training large language models (LLMs) from scratch \cite{bib17}. Given that we obtained approximately 50 million tokens for training, we decided to use GPT-2 small, which has 100 million parameters.

\subsubsection{Textualization of Indoor Location Information}\label{subsubsec1}
In the previous studies \cite{bib6,bib7}, outdoor location information has been encoded using regional grid codes defined by the Ministry of Internal Affairs and Communications of Japan. However, since grid codes are based on latitudes and longitudes, they are not suitable for application to indoor location information. Therefore, we introduced the following encoding method for use in retail stores, wherein the space is divided into six hierarchical levels, and each location is represented by six characters as depicted in Fig. 1(a).

Considering that Store A has dimensions of approximately 30m × 30m and the location resolution of the trajectory tracking system is approximately 50cm, we aim to achieve a grid size of 50cm × 50cm by dividing a space of 32m × 32m, which fully accommodates the store. For the first level, we divided the 32m × 32m space into four equal parts, assigning the characters “a,” “b,” “c,” and “d” to each respective section. Then, for the second level, we divided each of the four 16m × 16m squares into four equal parts, assigning the characters "e," "f," "g," and "h" to each respective section. We repeated this process six times to finally achieve a grid size of 50cm × 50cm and assigned characters "u," "v," "w," and "x" to each respective section, for the sixth level. It is important to note that "a" through "d" represent absolute coordinates, while "e" through "x" represent relative coordinates. Following these rules, all location information of a trajectory is represented by six-character strings, "agkpqw" for example. By arranging the strings representing the locations in chronological order, we create text describing the entire trajectory.

\subsubsection{Modeling of Purchasing Behavior}\label{subsubsec2}
Next, we modeled purchasing behaviors using the trajectories and matched retail scanner data. The term "purchase" here refers to the act of placing items into the basket in front of the product shelves. It should be noted that, since only the positions of products on regular shelves can be identified from the available layout diagrams, this study focuses solely on purchases of products sold on regular shelves. Products on end caps or seasonal displays are not considered.

First, we applied DBSCAN clustering \cite{bib10} to each trajectory in order to model pauses occurring in front of shelves. The DBSCAN clustering was performed on the x-y-time (xyt) space, with 1 second equivalent to 0.1 meters. The radius parameter $ 
\varepsilon$ was determined to be 2.5 meters, because at this value, the number of classified clusters became close to the actual number of purchased items in the given trajectory data. Fig. 1(b) shows an example of the clustering results. Three clusters are observed, with the first two likely indicating product examination in front of the shelves, while the last cluster represents waiting at the checkout.

Next, using the product categories of purchased items from retail scanner data and the store layout diagrams, we identified the zones where the items were placed as shown in Fig. 1(c) and (d). If a DBSCAN cluster exists within the identified zone, we consider the item to have been placed in the basket at the last point of the cluster, assuming that the decision to purchase was made at the last point after examining some items within the cluster. If no cluster exists within the identified product zone, we randomly select one point from the points passing through the zone and consider the item to have been placed in the basket at that point, assuming that the purchase was made casually without examining the items.. The estimated purchase positions obtained by this process are illustrated in Fig. 1(e). If no trajectory passing through the identified product zone is found, the item is excluded from the purchase list for the purpose of this study, considered as a purchase from seasonal displays or end caps.

Finally we represented the number of purchased items at each time point numerically and inserted it after the corresponding location characters in the encoded trajectories, followed by a period "." to indicate the end of the sentence. Fig. 1(f) shows the entire in-store journey represented by the encoding method.

\subsubsection{Tokenization and Training}\label{subsubsec3}
The text representing the chronology of locations and purchases created in the previous step was tokenized using Byte-level Byte Pair Encoding \cite{bib11}. The text is divided into byte levels, and frequently appearing bytes are combined repeatedly until the specified vocabulary size is reached. We set the vocabulary size to 50,000.

The tokenized result was used to train the GPT-2 small architecture from scratch. Training utilized 64\% of the entire data, with 16\% for validation and 20\% for testing. We adopted the Adam optimizer with a learning rate of 3e-4 and $\varepsilon$ set to 1e-08. The training process concluded after 10 epochs. 

\subsection{Fine Tuning the Pre-trained Model with Other Store Data}\label{subsec2}
Based on the pre-trained model trained on data from Store A, fine-tuning was performed using data from Store B \cite{bib12}. Utilizing the same encoding method as Store A, the location information for Store B was represented with six characters in the same manner. The purchasing behaviors were also converted into text similarly to Store A. The resulting text was tokenized and used as input to perform additional training on the pre-trained model. 

The numbers of training data from Store B used for fine-tuning ranged from 8 to the entire data  (approximately 20,000 records). To investigate how much training data can be reduced through fine-tuning, we perform generation using separate models for different numbers of training data. We adopted the Adam optimizer with a learning rate of 3e-4 and $\varepsilon$ set to 1e-08. Training concluded after 3 epochs.

\section{Results}\label{sec4}
\subsection{Generation Accuracy of the Pre-trained Model}\label{subsec3}
For each in-store customer journey classified as test data, the initial 7 time points, equivalent to the first 30 seconds of location and purchase information, were inputted into the pre-training model, followed by the generation of subsequent journey. An example of the generated results is shown in Fig. 2. After entering the store, the customer walks straight along the aisle, then circles around the perimeter and examines the products in the zone on the far right. After putting one item into the basket, he enters the inner aisle on his way to the checkout, where he puts two more items into the basket before heading to the register.

\subsubsection*{Trajectories}
To assess the accuracy of the generated results, we created a heatmap of customer traffic volume from generated results and compared it with the one created from actual data as shown in Fig. 3. The heatmaps were both created using the same 2000 data samples, randomly selected from the entire test data. Simply comparing them, it's evident that there's a similarity in the distribution of customer traffic within the store between the actual data and the generated results.

Additionally, to verify the accuracy of our proposed method, we generated customer trajectories using the method from a previous study \cite{bib14} and compared the results. Specifically, following the best parameters from the previous study, we trained an LSTM model with the past seven time points of $x$ and $y$ coordinates, as well as the calculated velocity $v$ and movement direction $\alpha$ derived from the history of the $x$ and $y$ coordinates, to predict the next $x$ and $y$ coordinates. This process was repeated recursively to generate 10 minutes (120 time points) of $x$ and $y$ coordinates, as shown in Fig. 4. The resulting trajectories include paths that cross over shelves and extend beyond the store boundaries, indicating that while the method from the previous study is very good at predicting the next single position, it is not suitable for generating entire in-store customer journeys sequentially, which is our current purpose.

\subsubsection*{Purchase Behaviors}
We aggregated the number of purchases in each zone and compared it with the actual data as shown in Fig. 5. The zones are sorted in descending order of purchase quantity for better readability. From the figure, while there are some fluctuations, it is evident that in zones with many purchases, a large number of purchase behaviors are generated, while in zones with fewer purchases, fewer purchase behaviors are generated, indicating that the pre-trained model generally simulates in-store customer journeys in store A well. To quantify this, we computed the JS divergence between the test data and the generated results. The JS divergence $D_{JS}(p,q)$ is defined by the following equation and quantitatively evaluates the similarity between two probability distributions $p(x)$ and $q(x)$. When the two distributions are perfectly aligned, $D_{JS}$ takes the value of 1, and it approaches 0 when there is no alignment at all.

\begin{equation}
D_{JS}(p,q) = \frac{1}{2}D_{KL}(p,M) + \frac{1}{2}D_{KL}(q,M)\label{eq1}
\end{equation}
where,
\begin{equation}
M(x) =  \frac{1}{2}p(x) + \frac{1}{2}q(x)\label{eq2}
\end{equation}
\begin{equation}
D_{KL}(p,q) =  \Sigma{p(x)\log{\frac{p(x)}{q(x)}}}\label{eq3}
\end{equation}
\\
The JS divergence between the test data and the generated results was 0.00973.

To validate the accuracy of our proposed method, we employed a previous study's approach \cite{bib16} to predict purchase behaviors and compared the results. Specifically, following the best parameters from the previous study, we constructed SVM models with an RBF kernel ($\sigma=0.4$) for each zone, using time spent in the zone and the customer's age as features to predict whether a purchase was made. Then, using the actual customer trajectories and ages, we calculated the time spent in each zone and input these values into the respective zone prediction models to predict purchases. The aggregated results are illustrated in Fig. 6. For comparison, we also included the aggregated purchase results generated simultaneously with the trajectories by our model. Since the previous study's model predicts whether a purchase is made rather than the number of purchases, the vertical axis in the figure represents the probability of making at least one purchase per visit in that zone, rather than the number of purchases per visit as in Fig. 5. As illustrated, our proposed model predicts purchases with higher accuracy in most zones. Moreover, while the SVM model shows significant deviations in some zones, our proposed model does not exhibit such large discrepancies in any zone. When calculating the JS divergence with the actual data, our proposed model achieves a value of 0.0100, while the previous study's model results in 0.0931, demonstrating that our model predicts purchases more accurately.

\subsection{Generation Accuracy of Fine Tuned Model with Less Data}\label{subsec4}
In Fig. 7, we compared the learning curves between the model fine-tuned with data from Store B using a pre-trained model from Store A and the model trained from scratch using data from Store B alone. The vertical axis represents the cross-entropy loss in the figure. It can be observed that fine-tuning with approximately 100 data samples yielded comparable accuracy to using around 20,000 data samples without fine-tuning.
We investigated the reproducibility of purchase quantity distribution generated by the fine-tuned models with different number of training data. The aggregated purchase quantities by zone for each model is shown in Fig. 8. This figure is made from the results of generating 8,000 trajectories for each model. It can be observed that as the number of training data increases, the deviation from the test data decreases. To quantify this trend, we computed and compared the JS divergence between the test data and the generated results. The JS divergence was 0.0374 when the training data size was 2048, and it decreased to 0.0161 when the training data size was 16384.

\section{Conclusion}\label{sec5}
We successfully applied the Transformer-based human trajectory generation AI proposed by Mizuno et al. \cite{bib6} to indoor trajectories, and further incorporated modeling of purchasing behavior. In the results generated by the proposed method, both the heatmap of in-store traffic and the distribution of purchases by zone were well reproduced.

On the other hand, there is room for further improvement in the block segmentation used during the textualization of location information in our approach. Drawing from prior research that deals with movement between predefined blocks within a store, we aim to explore and develop more effective block segmentation methods in future work.

Furthermore, we explored the possibility of fine-tuning pre-trained models, initially trained on data from one store, using data from other stores. As a result, we found that fine-tuning with a small amount of data yielded similarly low losses as training from scratch using the entire dataset. Specifically, in this study, fine-tuning with a pre-trained model reduced the required training data to as little as one-hundredth.  This suggests the potential for significantly reducing the amount of training data required for generating in-store customer journeys in other stores, thereby implying the possibility of significantly reducing data acquisition costs.

In the future, we aim to achieve higher precision generation with much less data by inputting background information into the model as special tokens. Possible background information includes customer attributes such as gender and age group, as well as environmental information such as store layout, weather conditions, and in-store congestion. Alternatively, by leveraging ideas similar to those used in natural language translation, it may be possible to develop new "translation" technology where, for example, inputting today's shopping purpose would generate trajectories tailored to that purpose. Furthermore, based on a multimodal approach, it may be possible to output trajectories from virtual store layout images, for example. This study will serve as a foundation for such future endeavors.


\section*{Acknowlegement}
We would like to express our gratitude to TSURUHA HOLDINGS, INC.
for providing the valuable data and engaging in regular discussions. This work was supported by JST CREST Grant Number JPMJCR20D3 and JSPS KAKENHI Grant Numbers JP21H01569, JP18KK0316.

\bibliography{references}% common bib file

\begin{thebibliography}{18}
\providecommand{\natexlab}[1]{#1}
\providecommand{\url}[1]{{#1}}
\providecommand{\urlprefix}{URL }
\providecommand{\doi}[1]{\url{https://doi.org/#1}}
\providecommand{\eprint}[2][]{\url{#2}}
 \bibcommenthead

\bibitem[{Das et~al(2020)Das, Kolvig-Raun, and Kj\ae{}rgaard}]{bib4}
Das A, Kolvig-Raun ES, Kj\ae{}rgaard MB (2020) Accurate trajectory prediction in a smart building using recurrent neural networks. In: Adjunct Proceedings of the 2020 ACM International Joint Conference on Pervasive and Ubiquitous Computing and Proceedings of the 2020 ACM International Symposium on Wearable Computers. Association for Computing Machinery, New York, NY, USA, UbiComp/ISWC '20 Adjunct, p 619–628, \doi{10.1145/3410530.3414319}, \urlprefix\url{https://doi.org/10.1145/3410530.3414319}

\bibitem[{Ester et~al(1996)Ester, Kriegel, Sander, and Xu}]{bib10}
Ester M, Kriegel HP, Sander J, et~al (1996) A density-based algorithm for discovering clusters in large spatial databases with noise. In: Proceedings of the Second International Conference on Knowledge Discovery and Data Mining. AAAI Press, KDD'96, p 226–231, \doi{10.5555/3001460.3001507}

\bibitem[{Hoffmann et~al(2024)Hoffmann, Borgeaud, Mensch, Buchatskaya, Cai, Rutherford, de~Las~Casas, Hendricks, Welbl, Clark, Hennigan, Noland, Millican, van~den Driessche, Damoc, Guy, Osindero, Simonyan, Elsen, Vinyals, Rae, and Sifre}]{bib17}
Hoffmann J, Borgeaud S, Mensch A, et~al (2024) Training compute-optimal large language models. In: Proceedings of the 36th International Conference on Neural Information Processing Systems. Curran Associates Inc., Red Hook, NY, USA, NIPS '22

\bibitem[{Horikomi et~al(2023)Horikomi, Fujimoto, Ishikawa, and Mizuno}]{bib7}
Horikomi T, Fujimoto S, Ishikawa A, et~al (2023) Generating individual trajectories using gpt-2 trained from scratch on encoded spatiotemporal data. arXiv preprint arXiv:230807940

\bibitem[{HuggingFace(year not available)}]{bib12}
HuggingFace (year not available) Fine-tune a pretrained model. \urlprefix\url{https://huggingface.co/docs/transformers/en/training}, accessed on July 8, 2024

\bibitem[{Ishimaru et~al(2021)Ishimaru, Morita, and Goto}]{bib3}
Ishimaru Y, Morita H, Goto Y (2021) In-store journey model with purchasing behavior based on in-store journey data and id-pos data. The Review of Socionetwork Strategies 15:215--237. \doi{10.1007/s12626-021-00078-5}

\bibitem[{Mizuno et~al(2022)Mizuno, Fujimoto, and Ishikawa}]{bib6}
Mizuno T, Fujimoto S, Ishikawa A (2022) Generation of individual daily trajectories by gpt-2. Frontiers in Physics 10. \doi{10.3389/fphy.2022.1021176}, \urlprefix\url{https://www.frontiersin.org/articles/10.3389/fphy.2022.1021176}

\bibitem[{Quuppa(2024)}]{bib8}
Quuppa (2024) Quuppa intelligent locating system. \urlprefix\url{http://quuppa.com/}, accessed on Mar. 31, 2024

\bibitem[{Radford et~al(2019)Radford, Wu, Child, Luan, Amodei, and Sutskever}]{bib9}
Radford A, Wu J, Child R, et~al (2019) Language models are unsupervised multitask learners. In: OpenAI blog

\bibitem[{Roscher et~al(2020)Roscher, Bohn, Duarte, and Garcke}]{bib18}
Roscher R, Bohn B, Duarte MF, et~al (2020) Explainable machine learning for scientific insights and discoveries. IEEE Access 8:42200--42216. \doi{10.1109/ACCESS.2020.2976199}

\bibitem[{Terano et~al(2009)Terano, Kishimoto, Takahashi, Yamada, and Takahashi}]{bib2}
Terano T, Kishimoto A, Takahashi T, et~al (2009) Agent-based in-store simulator for analyzing customer behaviors in a super-market. In: Vel{\'a}squez JD, R{\'i}os SA, Howlett RJ, et~al (eds) Knowledge-Based and Intelligent Information and Engineering Systems. Springer Berlin Heidelberg, Berlin, Heidelberg, pp 244--251, \doi{10.1007/978-3-642-04592-9_31}

\bibitem[{Tsai et~al(2017)Tsai, Li, and Kuo}]{bib14}
Tsai CY, Li MH, Kuo R (2017) A shopping behavior prediction system: Considering moving patterns and product characteristics. Computers \& Industrial Engineering 106:192--204. \doi{https://doi.org/10.1016/j.cie.2017.02.004}, \urlprefix\url{https://www.sciencedirect.com/science/article/pii/S0360835217300517}

\bibitem[{Wang et~al(2020)Wang, Cho, and Gu}]{bib11}
Wang C, Cho K, Gu J (2020) Neural machine translation with byte-level subwords. Proceedings of the AAAI Conference on Artificial Intelligence 34(05):9154--9160. \doi{10.1609/aaai.v34i05.6451}, \urlprefix\url{https://ojs.aaai.org/index.php/AAAI/article/view/6451}

\bibitem[{Wang et~al(2019)Wang, Wang, Zhang, Lu, and Wu}]{bib5}
Wang P, Wang H, Zhang H, et~al (2019) A hybrid markov and lstm model for indoor location prediction. IEEE Access 7:185928--185940. \doi{10.1109/ACCESS.2019.2961559}

\bibitem[{Wang et~al(2022)Wang, Yang, and Zhang}]{bib1}
Wang P, Yang J, Zhang J (2022) A spatial-contextual indoor trajectory prediction approach via hidden markov models. Wireless Communications and Mobile Computing \doi{10.1155/2022/6719514}

\bibitem[{Zhao et~al(2021)Zhao, Zuo, Zhao, and Jiang}]{bib13}
Zhao W, Zuo Y, Zhao L, et~al (2021) Application of lstm models to predict in-store trajectory of customers. 2021 International Conference on Data Mining Workshops (ICDMW) pp 288--294. \urlprefix\url{https://api.semanticscholar.org/CorpusID:246080538}

\bibitem[{Zuo et~al(2014)Zuo, Ali, and Yada}]{bib16}
Zuo Y, Ali AS, Yada K (2014) Consumer purchasing behavior extraction using statistical learning theory. Procedia Computer Science 35:1464--1473. \doi{https://doi.org/10.1016/j.procs.2014.08.209}, \urlprefix\url{https://www.sciencedirect.com/science/article/pii/S1877050914011740}, knowledge-Based and Intelligent Information \& Engineering Systems 18th Annual Conference, KES-2014 Gdynia, Poland, September 2014 Proceedings

\bibitem[{Zuo et~al(2016)Zuo, Yada, and Ali}]{bib15}
Zuo Y, Yada K, Ali AS (2016) Prediction of consumer purchasing in a grocery store using machine learning techniques. In: 2016 3rd Asia-Pacific World Congress on Computer Science and Engineering (APWC on CSE), pp 18--25, \doi{10.1109/APWC-on-CSE.2016.015}

\end{thebibliography}

\begin{figure}[htbp]
\centering
\includegraphics[width=1\textwidth]{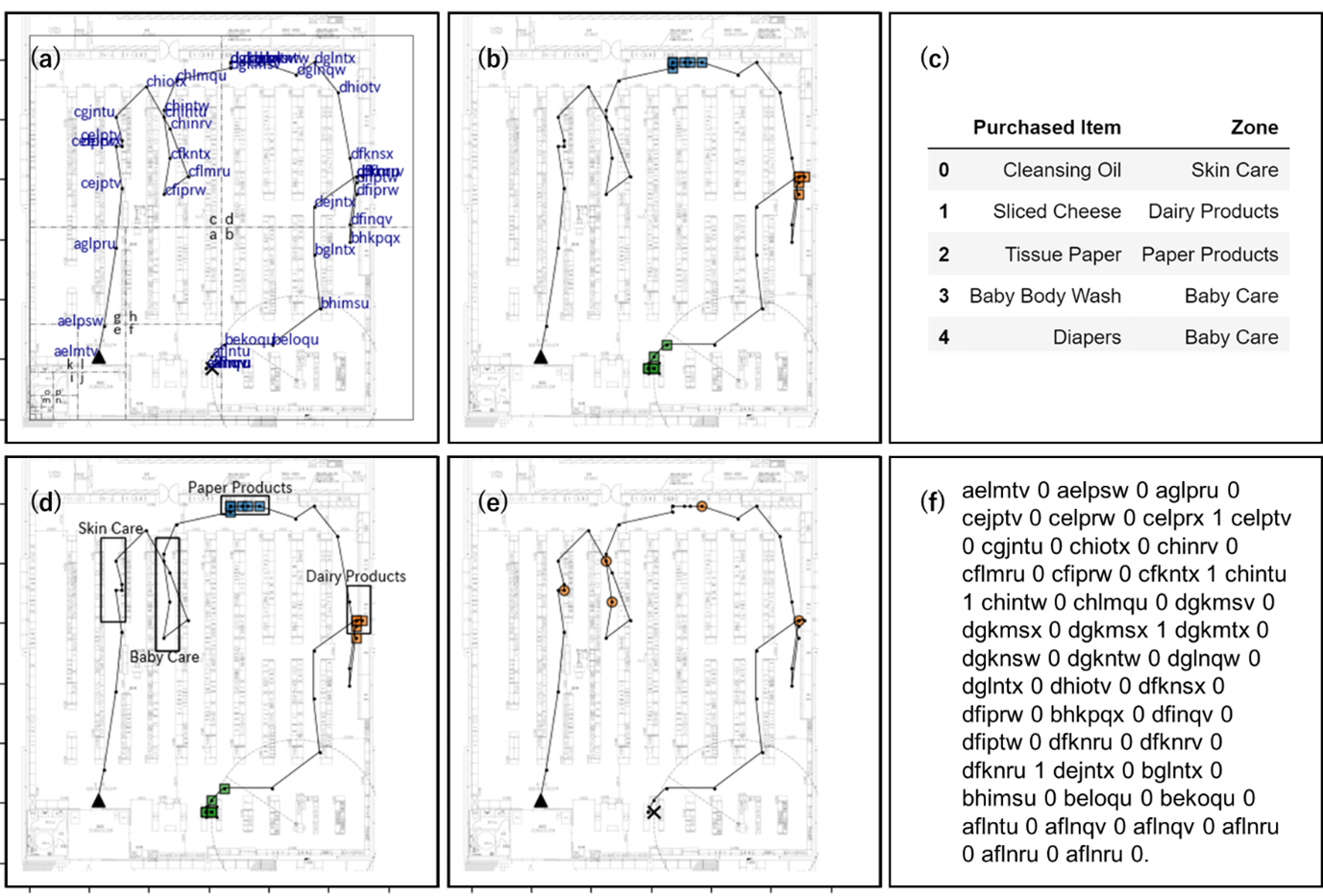}
\caption{Textualization of Location Information and Modeling of Purchase Behavior. (a) Division of the entire store into 50cm-grid meshes. $\blacktriangle$ denotes the starting point (entrance), × denotes the ending point (cash register). (b) DBSCAN clustering. blacksquare denotes a location belonging to a cluster. (c) (d) Identification of zones in which the purchased items were placed using the layout diagram. The rectangles surrounding the trajectories represent zones. (e) Estimation of purchase location. $\bullet$ denotes a location where a purchase action was taken. (f) Textualization of the whole in-store customer journey.}\label{fig1}
\end{figure}

\begin{figure}[htbp]
\centering
\includegraphics[width=0.7\textwidth]{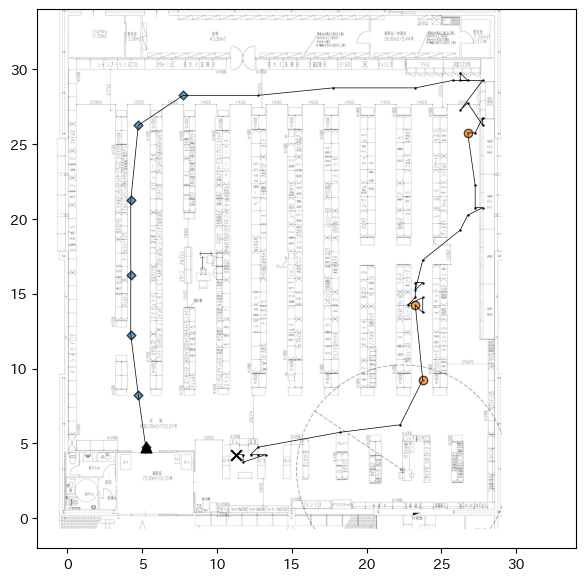}
\caption{Example of Generated Trajectory with Purchase Behavior: $\blacktriangle$ denotes the starting point (entrance), $\blacklozenge$ denotes a inputted location, "." denotes a generated location, $\bullet$ denotes a purchase, and × denotes the ending point (checkout register).}\label{fig2}
\end{figure}

\begin{figure}[htbp]
\centering
\includegraphics[width=0.9\textwidth]{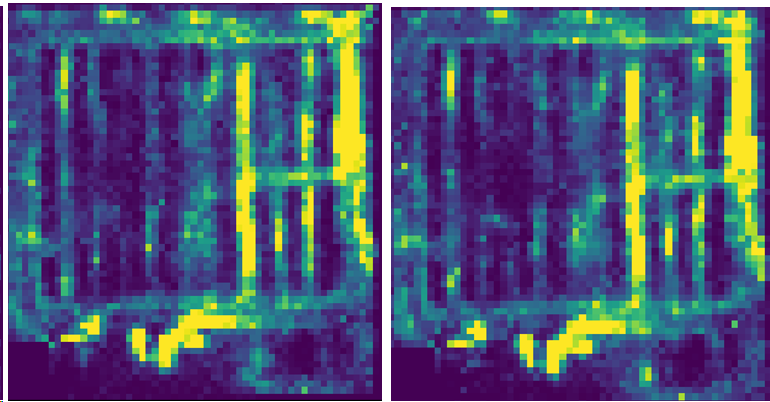}
\caption{Comparison of In-Store Traffic Heatmaps  (Left: Actual Data, Right: Generated Results).}\label{fig3}
\end{figure}

\begin{figure}[htbp]
\centering
\includegraphics[width=0.7\textwidth]{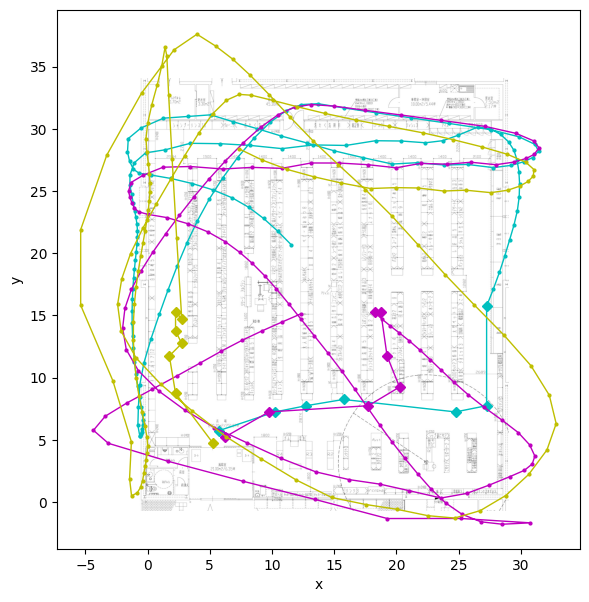}
\caption{Examples of three customer trajectories generated by LSTM. $\blacklozenge$ denotes a inputted location, "." denotes a generated location.}\label{fig4}
\end{figure}

\begin{figure}[htbp]
\centering
\includegraphics[width=1\textwidth]{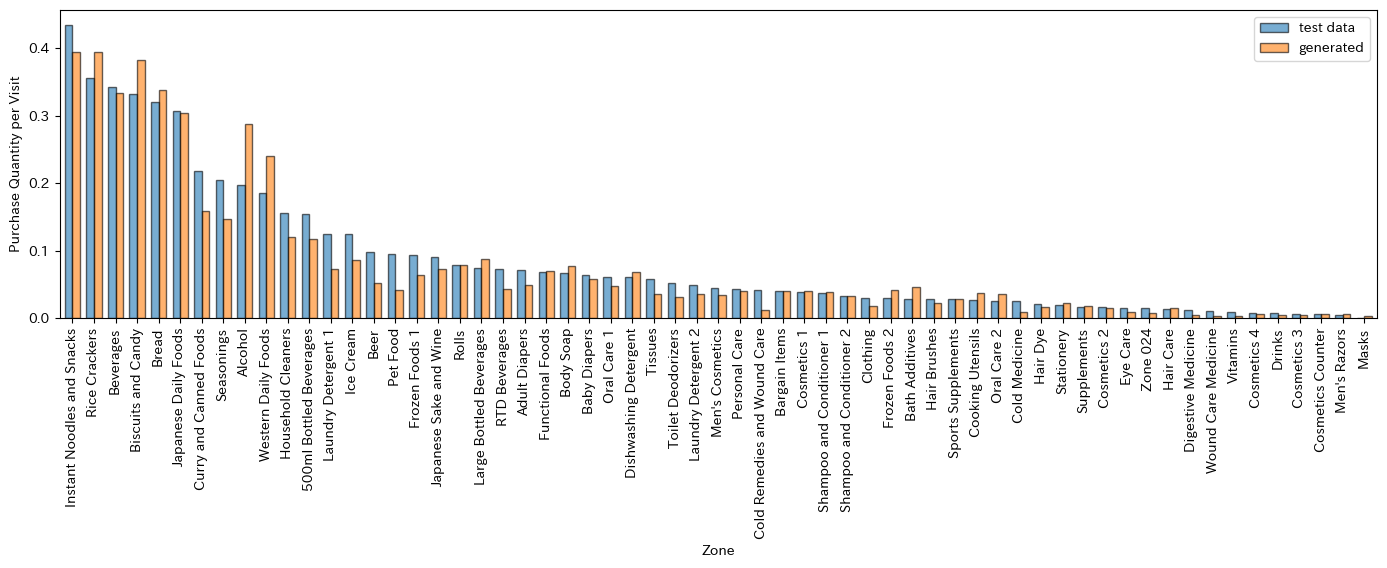}
\caption{Comparison of Purchase Counts by Zone. The numbers of purchased items per visit in each of the 61 zones in Store A are compared between the test data and the generated results.}\label{fig5}
\end{figure}

\begin{figure}[htbp]
\centering
\includegraphics[width=1\textwidth]{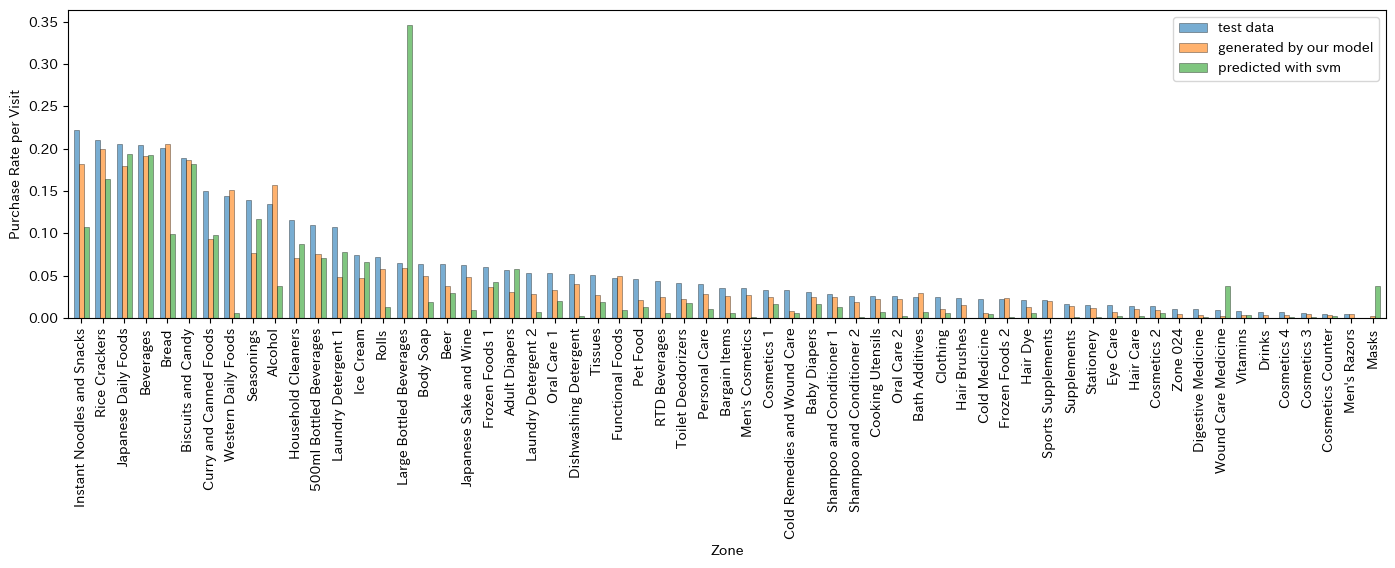}
\caption{Comparison of Purchase Occurrence by Zone with a Previous Study's Method.}\label{fig6}
\end{figure}

\begin{figure}[htbp]
\centering
\includegraphics{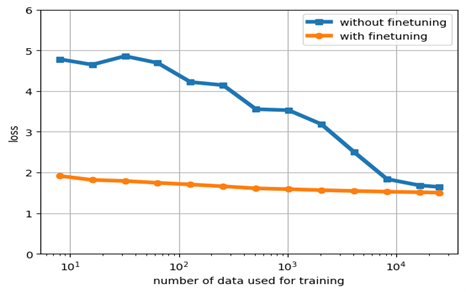}
\caption{The learning curves with and without fine-tuning.}\label{fig7}
\end{figure}

\begin{figure}[htbp]
\centering
\includegraphics[width=1\textwidth]{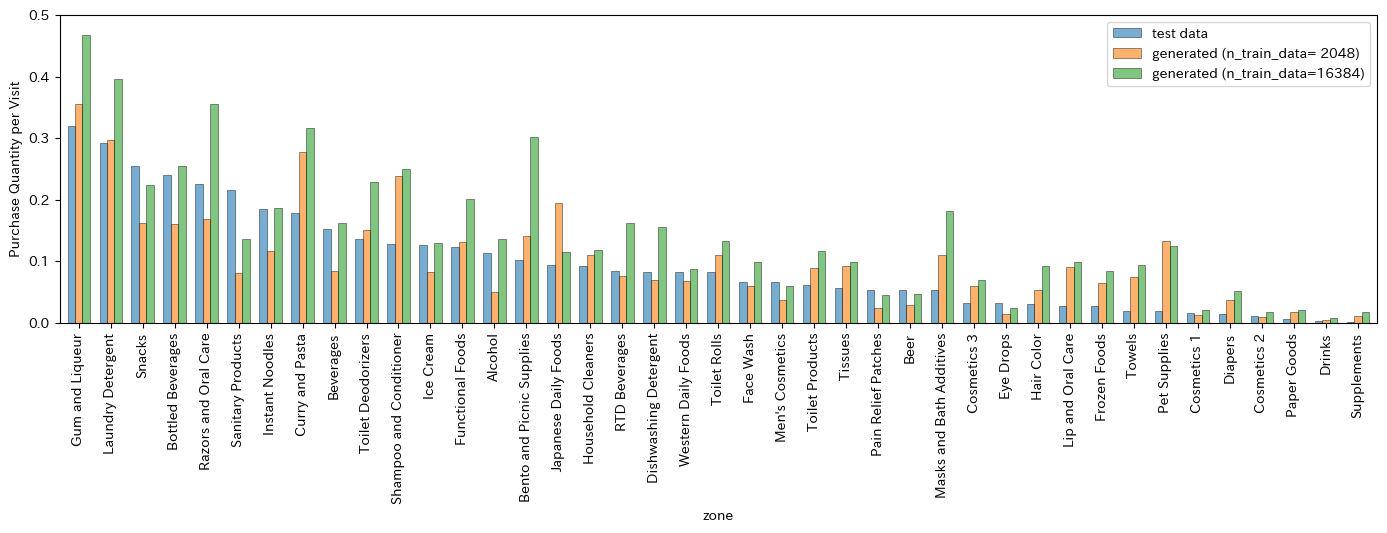}
\caption{Comparison of purchase quantities by zone generated by the fine-tuned models with different number of training data.}\label{fig8}
\end{figure}

\end{document}